\documentclass[conference]{IEEEtran}
\IEEEoverridecommandlockouts
\usepackage{cite}
\usepackage{amsmath,amssymb,amsfonts}
\usepackage{algorithmic}
\usepackage{graphicx}
\usepackage{textcomp}
\usepackage{xcolor}
\usepackage{hyperref}

\newcommand{\norm}[1]{\left\lVert#1\right\rVert}

\newcommand{\mean}[1]{\mathbb E#1}

\newcommand{\trace}[1]{\text{Tr}\left[#1\right]}
\newcommand{\dd}[1]{\mathrm{d}#1}
\newcommand{\traceLim}[2][17.5]{ \text{Tr}_{#1}\left[#2\right] }

\def\BibTeX{{\rm B\kern-.05em{\sc i\kern-.025em b}\kern-.08em
    T\kern-.1667em\lower.7ex\hbox{E}\kern-.125emX}}
\begin{document}

\title{Gradient flow on extensive-rank positive semi-definite matrix denoising}

\author{\IEEEauthorblockN{Antoine Bodin and Nicolas Macris}
\IEEEauthorblockA{
School of Computer and Communication Sciences\\
SMILS - IC - Ecole Polytechnique F\'ed\'erale de Lausanne\\
antoine.bodin@epfl.ch and nicolas.macris@epfl.ch}
}

\maketitle

\begin{abstract}
In this work, we present a new approach to analyze the gradient flow for a positive semi-definite matrix denoising problem in an extensive-rank and high-dimensional regime. We use recent linear pencil techniques of random matrix theory to derive fixed point equations which track the complete time evolution of the matrix-mean-square-error of the problem. The predictions of the resulting fixed point equations are validated by numerical experiments. In this short note we briefly illustrate a few predictions of our formalism by way of examples, and in particular we uncover continuous phase transitions in the extensive-rank and high-dimensional regime, which connect to the classical phase transitions of the low-rank problem in the appropriate limit. The formalism has much wider applicability than shown in this communication.
\end{abstract}


\section{Introduction}

Matrix denoising and factorization play a crucial role in a variety of data science tasks such as matrix sensing, phase retrieval or synchronisation, or matrix completion. The problem consists in reducing the amount of noise or irrelevant information present in a dataset, allowing for more accurate analysis and interpretation of the data, as well as better computational efficiency and modeling by way of dimensionality reduction. The literature on the subject is immense and we refer to \cite{ChenChi, chi2019nonconvex} for recent overviews of applications and theory in various settings and formulations.

In this contribution we focus on the study of gradient-flow for the following statistical formulation for positive definite matrix denoising. We consider a "ground truth" signal $X^* \in \mathbb R^{n \times d}$ with randomly sampled independent entries $X^*_{ij} \sim \mathcal N(0,\frac{1}{n})$ where the dimensions $n,d$ are  such that $\phi = \frac{d}{n}$ is fixed. Then we define the corrupted data matrix $Y \in \mathbb R^{n \times n}$
\begin{equation}\label{model1}
    Y = X^* X^{*T} + \frac{1}{\sqrt{\lambda}} \xi
\end{equation}
where $\xi$ is an additive symmetric random noise with $\xi_{ij} = \xi_{ji} \sim \mathcal N(0,\frac{1}{n})$ and $\lambda$ is (proportional to) the signal-to-noise ratio. The objective is to estimate the ground truth positive semi-definite matrix $X^* X^{*T}$ from the corrupted data matrix $Y$ with a matrix $XX^T$ such that $X \in \mathbb R^{n \times m}$ where $m$ is set from the fixed ratio $\psi = \frac{m}{n}$. Note that we allow $d$ and $m$ to be different. 
The estimator studied in this contribution is given by the gradient flow $X(t)$ ($t$ is  time) for an objective function with regularization parameter $\mu$, defined as 
\begin{equation}
    \mathcal H(X) = \frac{1}{4 d} \norm{Y - X X^{T}}_{F}^{2} + \frac{\mu}{2d} \norm{X}_{F}^{2} 
\end{equation}
where $\norm{ \cdot }_F$ is the Frobenius norm. The initialization of gradient flow is $X(0)= X_0 \in \mathbb R^{n \times m}$ random with i.i.d matrix elements from  
$\mathcal N(0,\frac{1}{n})$.
As a measure of performance we adopt the expected matrix-mean-square-error 
\begin{equation}
\mathbb{E}\mathcal E = \frac{1}{d} \mathbb{E}\norm{X^* X^{*T} - X X^{T}}_{F}^{2} 
\end{equation}
where the expectation is over $\xi$, $X^*$, $X_0$.
Note that the objective function and performance measure are not the same and can be thought of as "training" and "generalization" errors in the language of machine learning.

{\bf  Summary of main contributions}: 
\begin{itemize}
\item
We derive a set of analytical fixed point equations whose solutions allow to compute the full performance curve $t \to \mathbb{E}\mathcal{E}_t$ 
for the extensive-rank and high-dimensional regime where $m, d, n$ all tend to infinity while $\phi, \psi$ are kept fixed (results 1 and 2 in Sec. \ref{sec:results}). Continuous time average behaviour of gradient flow is a proxy for the usual discrete gradient descent algorithm, and has the advantage that it is more amenable to analytical study. The numerical experiments confirm that (a) $\mathcal{E}_t$ concentrates over its expectation; (b) theoretical predictions of gradient flow agree with gradient descent. See Fig. \ref{fig:graph1}.
\item
We further push the analysis of these equations in the time limit $t=+\infty$ and display specific examples where a critical value $\lambda_c$ can be calculated such that: (a) for $\lambda\leq\lambda_c$ the performance error of gradient flow is no better than the one of the null-estimator $X=0$; (b) for $\lambda>\lambda_c$ better estimation is possible; (c) the phase transition between the two regimes is a continuous type phase transition. These results are displayed on Fig. \ref{fig:graph3}.
\item
We analyze the limit $\phi = \psi \to 0$ (after $n, m, d$ have been sent to infinity) and derive a connection with the low-rank setting. It turns out that the matrix-mean-square-error curve (when $t\to +\infty$) tends to the one of the rank-one problem and the phase transition reduces to the well known BBP transition at $\lambda_c=1$. 
\end{itemize}
 
We use tools based on modern results in random matrix theory. Central to our derivations, is the formalism of \emph{linear-pencils}, that initially appeared in \cite{spectra, mingo2017free} and has been further improved recently in the context of neural networks \cite{adlam2020neural, bodin2021model, bodin.2212.06757}. In particular we make use of extensions provided in \cite{bodin.2212.06757} to derive closed-form expressions of non-trivial averages over $\xi$, $X^*$, $X_0$ appearing in traces of complicated "rational" expressions of these random matrices. Although these techniques have not yet always been rigorously proven they have been used successfully in different applications, and the predictions are confirmed by numerical experiments. In addition, we use holomorphic functional calculus for matrices \cite{schwartz1958linear}.

{\bf  Brief review of literature}:
The full time-evolution of gradient flow for the rank-one problem (the so-called spiked Wigner model with $d=m=1$) has been solved and rigorously analyzed in much the same spirit as the present work in \cite{pmlr-v134-bodin21a} with the difference that the spike is constrained to lie on a sphere all along the evolution. 
For the present extensive-rank setting rigorous or even analytical results on the whole time-evolution are scarce. Closely connected to our work is the recent paper \cite{tarmoun2021understanding}. An essential difference however is that in \cite{tarmoun2021understanding} the initialization $X(0)= X_0$ is taken to have eigenvectors aligned with those of $Y$ (this pre-processing can be implemented empirically in practice). Moreover the authors do not carry out the random matrix averages fully analytically. Gradient flow has been studied in a variety of settings more or less related to the present one, see  \cite{gunasekar2017implicit, chou2020gradient, saxe2013exact, mannelli2019passed, BenArous-et-al-2022, Liang-Sen-Sur-2022}.


Bayesian approaches are quite well understood for the low-rank problem (mainly rank-one). This context is quite different from the present one. To begin with it is not dynamical. One studies the Minimum-Mean-Square-Estimator (MMSE) computed as the conditional expectation of the signal with respect to the Bayesian posterior probability distribution \cite{Montanari-Richard-2014, lelarge2019fundamental, luneau2020high, barbier2019adaptive, miolane2017fundamental, pourkamali2021mismatched, pourkamali2022mismatched, CamilliContucciMingione, barbier2022price}. Bayesian-optimal as well as mismatched estimation settings have been well studied with rigorous results on the mutual information, the MMSE, the cross-entropy, and the problem displays a rich phenomenology of first and higher order phase transitions depending on the nature of the priors. Related dynamics of the Approximate Message Passing (AMP) algorithms is also well understood for these problems \cite{DBLP:conf/isit/LesieurMLKZ17, Lesieur_2017, Montanari2017EstimationOL} . The realm of extensive-rank within such Bayesian and AMP approaches is quite open and very timely \cite{Kabashima_2016, barbier2021statistical, maillard2021perturbative, troiani2022optimal, camilli2022matrix}.

Finally other types of non-dynamical approach belong to the class of spectral methods like Principal Component Analysis (PCA).  The low rank case is covered by \cite{baik2005phase, Pch2004TheLE, benaych2011eigenvalues}. 
For the extensive-rank setting the results are scarce and little is known except for ensembles of rotation invariant signals for which an interesting class of Rotation Invariant Estimators (RIE) has been proposed \cite{bun2017cleaning}.

\section{Results}\label{sec:results}

\subsection{Preliminaries}
We simplify the notations by introducing the variables $Z = X X^T$ and $Z^* = X^* X^{*T}$ and the \emph{order parameters} $p$ and $q$ such that $\mathcal E = r - 2 q + p$ with:
\begin{equation}
    q = \frac{1}{d} \trace{Z^*  Z} \quad \quad p = \frac{1}{d} \trace{Z^2} \quad  \quad r = \frac{1}{d} \trace{(Z^*)^2}
\end{equation}
In the rank-one setting, $p$ can be seen as a norm of the estimator while $q$ represents the angle with the ground-truth.
We consider the gradient flow
\begin{equation}
\frac{\dd X_t}{\dd t} = -\phi \nabla \mathcal H(X_t)
\end{equation}
and track the evolution of the matrix mean-square error $\mathcal E_t$ through the quantities $q_t$ and $p_t$. The factor $\phi$ amounts to a rescaling of time which leads to more convenient expressions.
With the additional notation $H = Y - \mu I_n$, expanding the gradient provides: $\frac{\dd X_t}{\dd t} = (H-Z_t) X_t$, which in turns provides the matrix Riccati differential equation:
\begin{equation}
    \frac{\dd Z_t}{\dd t} = H Z_t + Z_t H - 2Z_t^2
\end{equation}
A general solution of this matrix differential equation is (see e.g., \cite{tarmoun2021understanding}):
\begin{equation}\label{eq:Zt}
    Z_t = e^{tH} X_0 \left(I_m + 2 X_0^T \int_0^t e^{2sH} \dd s X_0 \right)^{-1} X_0^T e^{tH}
\end{equation}
This formula is valid regardless of the dimensions $n,m,d$. In particular, when $m=d=1$ this is the solution of the rank-1 gradient flow.
In the high-rank case, it is not straightforward a priori how to track the evolution of the matrix $Z_t$ as firstly the rank of $X_0 X_0^T$ and $X^* X^{*T}$ (or $Y$ or $H$) are not necessarily equal when $d \neq m$, and secondly because the eigenvectors of the two matrices are not aligned at the initialization.

In the following, we will consider the high-dimensional limit $n,m,d \to \infty$ with $d/n$ and $m/n$ fixed and make the following assumptions:
\begin{itemize}
    \item The limits of traces $p_t = \frac{1}{d}{\rm Tr}[Z_t^2]$, $q_t= \frac{1}{d}{\rm Tr}[Z^*Z_t]$ (and $\mathcal{E}_t$) concentrate on their expectation, as well as related traces used in the \emph{linear-pencils}  method in Sec. \ref{sec:sketchproof}.
    \item We assume that $H$ has a limiting spectral distribution whose support can be enlaced in a finite contour $\Gamma \subset \mathbb C$.
\end{itemize}
To keep notations lighter we shall abusively denote by $p_t$, $q_t$, $\mathcal{E}_t$ their limiting deterministic values.

\subsection{Main results}

The MSE $\mathcal E_t$ of the problem is completely given by $q_t$, $p_t$ and the constant $r$ which in the high-dimensional limit is found to be $r = 1 + \phi$ from the second moment of the Marchenko-Pastur law \cite{marchenko1967distribution}. The main contribution of this paper is the self-consistent set of equations that fully track $q_t$ and $p_t$:

\textbf{(Result 1)} In the high dimensional limit, the overlap $q_t$ evolves according the integral:
\begin{equation}\label{eq:qt}
    q_t = \int_{\mathbb R} 
    \frac{z \rho_Q(z) \dd z}{
        1 - e^{-2tz} + z \tilde q_t e^{-2tz}
    }
\end{equation}
with the auxiliary function $\tilde q_t$ solution of the fixed-point equation:
\begin{equation}\label{eq:tilde_qt}
    \psi \tilde q_t = 1 + \int_{\mathbb R}  \frac{
        (1- e^{-2tz}) \rho_P(z) \dd z
    }{
        \frac{1}{\tilde q_t} (1 - e^{-2tz}) + z e^{-2tz}
    }
\end{equation}
and $\rho_P, \rho_Q$ are given by their inverse Stieltjes transforms $P(z), Q(z)$. These are the analytic solutions of the degree 3 polynomials such that $-zP(z) \to 1$ when $|z| \to \infty$ and $-zQ(z) \to 1$ when $|z| \to \infty$ where:
\begin{equation*}
    P^{3} + P^{2} \left( \lambda (\mu + z) + 1\right) + P \lambda \left(  \mu + z -  \phi + 1\right) + \lambda = 0
\end{equation*}
\begin{equation}\label{eq:Q_and_P}
    Q^{3} \phi 
    +
    Q^{2} \left(\mu + z - 2 \phi - 1 - \frac{1}{\lambda}\right)
    - Q \left( \mu + z - \phi - 2 \right) = 1 
\end{equation}

\textbf{(Result 2)} In the high-dimensional limit, the eigenvalue distribution of $Z_t$ is found by the inverse Stieltjes-Transform of $h_t(z)$ where:
\begin{align}\label{eq:pt}
    h_t(z) & = \frac{-1}{2\pi i} \oint_{\Gamma}
    -\frac{
        (1 + e^{-2tx} (\frac{x}{\tilde h_t(z)} - 1)) P(x) \dd x
    }{
        x + z + z e^{-2tx} (\frac{x}{\tilde h_t(z)} -1)
    }  \\
    \tilde h_t(z) & = 1 + \frac{1}{\psi} \frac{-1}{2\pi i} \oint_{\Gamma}
    -\frac{
        (x + z - z e^{2tx}) P(x) \dd x
    }{
        x + z + z e^{-2tx} (\frac{x}{\tilde h_t(z)} -1)
    }
\end{align}
in particular, we find:
\begin{equation}\label{eq:derivative_pt}
    p_t = -\frac{1}{2\phi} \left.  \frac{\partial^{(2)}}{\partial z^2} \left( \frac{1}{z} h_t\left(\frac{1}{z}\right) \right) \right\rvert_{z=0}
\end{equation}
Note that a similar system of equations as \eqref{eq:qt} can be derived by calculating the first and second derivatives in $z$ as given by  \eqref{eq:derivative_pt} and using the integrands in \eqref{eq:pt}. However the resulting formulas are too cumbersome to be presented here.

\subsection{Discussions and experiments}

\begin{figure}[h]
    \centering
    \includegraphics[width=0.6\linewidth]{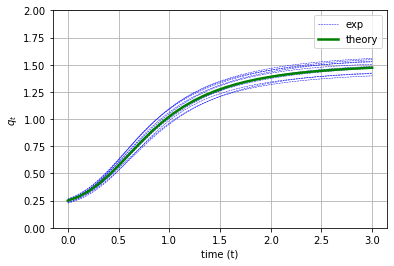}
    \vskip -0.2cm
    \caption{Comparison of $q_t$ evolution with $10$ runs of a gradient descent with $n=100$, $m=25$, $d=75$ and $\lambda=10^4$ and $\mu=0$. }
    \label{fig:graph1}
\end{figure}

Figure \ref{fig:graph1} provides an example of the calculation of $q_t$ through time compared with experimental runs: we see a good agreement between the curves and the prediction.

\textbf{Asymptotic Limit $t\to \infty$:}
An interesting question is to study the asymptotics of $q_t$ when $t \to \infty$. We take the ansatz that $q_t \sim \gamma e^{2 \alpha t}$ in this limit with $\alpha>0$ and $\gamma > 0$ another constant and plug this in equation \eqref{eq:tilde_qt}:
\begin{align}
    \psi & = \frac{1}{\tilde q_t} + \int_{\mathbb R}  \frac{
        (1- e^{-2tz}) \rho_P(z) \dd z
    }{
       1 - e^{-2tz} + z \tilde q_t e^{-2tz}
    } \\
    & \simeq \frac{e^{-2\alpha t}}{\gamma}
    + \int_{\mathbb R}  \frac{
        (1- e^{-2tz}) \rho_P(z) \dd z
    }{
       1 - e^{-2tz} + z \gamma e^{-2 (z-\alpha) t} 
    } \\
    & \simeq  \int_\alpha^\infty \rho_P(z) \dd z = 1-F_P(\alpha)
\end{align}
With $F_P$ the CDF of $P$. Such a solution exists when we can find $\alpha$ such that $F_P(\alpha) = 1 - \psi$, effectively selecting the proportion $\psi$ of the eigenvalues of $H$ in the interval $(\alpha, +\infty)$. Due to the assumption $\alpha >0$, a further condition for the existence of such an $\alpha$ is $F_P(0) < 1 - \psi \leq 1$ or: $0 \leq \psi < 1 - F_P(0) \leq 1$. This implies that the ansatz is valid in the \emph{under-parameterized} regime $(m<n)$. The asymptotic limit is thus given by
$q_\infty = \lim_{t \to \infty} q_t = \int_{\alpha}^\infty z \rho_Q(z) \dd z$.
Note that the alternative ansatz that $\tilde q_t$ converges towards a finite limit leads to a similar solution as but with $\alpha = 0$.

\begin{figure}[h]
    \centering
    \includegraphics[width=0.6\linewidth]{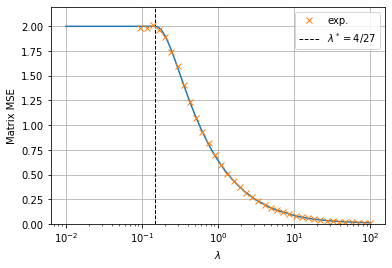}
    \vskip -0.3cm
    \caption{Experimental and theoretical $\mathcal E_\infty$ with $\phi=\psi=1, \mu = \frac{1}{\lambda}$ }
    \label{fig:graph3}
\end{figure}

A similar line of reasoning lead us to consider the term $p_\infty = \frac{1}{\phi} \int_{\alpha}^\infty z^2 \rho_P(z) \dd z$ and thus a asymptotic mean square error:
\begin{equation}\label{eq:mse_asymptotic}
    \mathcal E_\infty = r - \int_\alpha^\infty \left(2 z \rho_Q(z) - \frac{1}{\phi} z^2 \rho_P(z) \right) \dd z
\end{equation}
As an example, for $\phi = \psi = 1$ and $\mu = \frac{1}{\lambda}$, and $\alpha=0$ we expect from formula \eqref{eq:mse_asymptotic} that $\mathcal E_\infty = r$ when the support of $\rho_P$ and $\rho_Q$ is located below $0$. This can be found by studying the discriminant $\Delta_P(\lambda, z)$ of the polynomial solved by $P$: because it is a order $3$ polynomial with coefficients in $\mathbb R$ when $z \in \mathbb R$, either the solutions are all real $(\Delta_P > 0)$ implying $\rho_P(z) = 0$, or one is real and two are complex conjugate $(\Delta_P = 0)$ implying $\rho_P(z)>0$. At a specific $\lambda$, the support of $\rho_P$ is located below $0$ and touches $z=0$. This $\lambda_c$ is solution of $\Delta_P(\lambda_c, 0) = 0$ which provides the solution $\lambda_c = \frac{4}{27}$. The whole error curve at $t = +\infty$ is shown in Figure \ref{fig:graph3}.
The choice $\mu = \frac{1}{\lambda}$ is natural from a Bayesian point-of-view because it would correspond to the situation where the statistician matches its prior to the ground-truth when $\psi=\phi$.

\begin{figure}[h]
    \centering
    \includegraphics[width=0.9\linewidth]{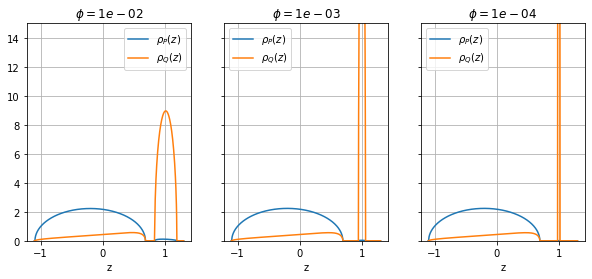}
    \vskip -0.3cm
    \caption{Bulk of eigenvalues for $\lambda = 5$ and different values of $\phi$}
    \label{fig:graph2}
\end{figure}

\begin{figure}[h]
    \centering
    \includegraphics[width=0.9\linewidth]{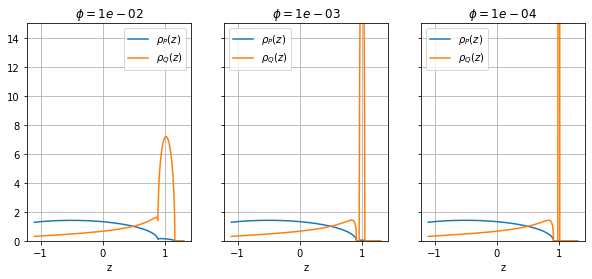}
    \vskip -0.3cm
    \caption{Bulk of eigenvalues for $\lambda = 2$ and different values of $\phi$}
    \label{fig:graph4}
\end{figure}

\textbf{Low rank limit when $\phi = \psi \to 0$:} We bring to the reader's attention that the objective function $\mathcal H$ when $d=1$ and $n\to \infty$ with $\mu = \frac{1}{\lambda}$ corresponds precisely to the spiked-Wigner problem. This suggest to look at the limit $\phi = \psi \to 0$. In this situation, we expect $\alpha$ should be close to the maximum eigenvalue of the bulk of $\rho_Q$. We make the following observation in Figure \ref{fig:graph2}: as $\phi$ decreases, $\rho_P$ in blue has two bulks of eigenvalues, one of which disappears as $\phi$ grows. On the other hand, $\rho_Q$ in orange displays also two bulks at the same locations but the second bulk develops a mass as $\phi \to 0$. Therefore, we expect that $\alpha$ adjusts itself to the maximum eigenvalue of the first bulk of $\rho_P$. Furthermore, interestingly we see that these two bulks are getting closer when $\lambda$ is closer to $1$ as seen in Figure \ref{fig:graph4}.

With these observations, we expect that $Q(z)$ has a pole $z = 1$ in the limit $\phi \to 0$. Let's consider a polynomial equation of $\hat Q$ solving the reduced polynomial equation of $Q$ with $\phi = 0$:
\begin{equation}
 \hat Q^{2} (z - 1) - \hat Q (z - 2 + \frac{1}{\lambda}) - 1 = 0 .
 \end{equation}
In order to find a potential pole, we consider $\mathcal Q(z) = (1-z) \hat Q(z)$ and check for potential limits of $\mathcal Q$ when $z\to 1$. First of all, injecting $\mathcal Q$ in the former polynomial equation, we find:
\begin{equation}
\left(\mathcal Q^{2} \lambda + \mathcal Q (\lambda z - 2 \lambda + 1) - \lambda z + \lambda\right)/(z-1) = 0 .
\end{equation}
Therefore, on the upper-complex plane we find the numerator equals $0$, and by analytic continuation, the limit $z \to 1$ follows: $\mathcal Q(1) (\mathcal Q(1) \lambda - \lambda +1) = 0$ so $\mathcal Q(1) \in \{ 0, 1 - \frac{1}{\lambda} \}$. It is interesting to notice the connection with the usual Bayesian overlap of the spiked Wigner model - since 
$\mathcal Q(1)$ represents the squared overlap $q_\infty$ in the limit $\phi\to 0$. Pushing further this analysis for $P(z)$ allows to eventually get $p_\infty$ and $\mathcal{E}_\infty$ in the limit $\phi\to 0$ and check the connection with the Bayesian MMSE of the spiked Wigner model. 


\section{Sketch of Proof}\label{sec:sketchproof}
Our method relies on considering the interaction of the random matrices $X_0, X^*, \xi$. We treat each term $q_t$ and $p_t$ separately with the linear-pencil technique. In both cases, we first factor out the $X_0$ matrix, then decouple the time dependency from the remaining random matrix expressions, and finally factor-out $X^*, \xi$.

Our results are derived in the limit $n,m,d\to +\infty$. For a sequence of matrices $A_N \in \mathbb R^{N\times N}$ we use the notation $\traceLim[N]{A_N} = \lim_{N \to \infty} \frac{1}{N}\trace{A_N}$. As stated in Sec. \ref{sec:results} we assume that the limiting traces involved in the linear pencil method concentrate.

\subsection{Tracking the angle $q_t$}
The term $q_t = \traceLim[d]{Z^* Z_t}$ can be completely recovered from a sub-block of the following linear-pencil $M_q$:
\begin{align}
    M_q = \left(
        \begin{array}{c|||c||c|ccc|c}
            0 & I_d & 0 & 0 & 0 & 0 & 0\\ \hline \hline  \hline
            I_d & 0 & 0 & 0 & 0 & 0 & W_t\\ \hline \hline
            0 & 0 & 0 & X_0 & 0 & 0 & I_n\\ \hline
            0 & 0 & X_0^T & I_m & 0 & X_0^T & 0\\
            0 & 0 & 0 & 0 & L_t & I_n & 0\\
            0 & 0 & 0 & X_0 & I_n & 0  & 0\\ \hline
            0 & W_t^T & I_n& 0 & 0 & 0 & 0
        \end{array}
    \right)
\end{align}
Where $W_t = X^{*T} e^{tH}$ and $L_t = 2 \int_0^t e^{2sH} \dd s$. A recursive application of the Schur-complement to compute $M_q^{-1}$ shows that the block $(M_q^{-1})^{(1,1)}$ is the random matrix $X^{*T} Z_t X^*$. So in fact: $q_t = \traceLim[d]{(M_q^{-1})^{(1,1)}}$.

The random matrices $X_0, X^*, \xi$ are all independent and $X_0$ is not part of the terms $W_t, L_t$. Therefore, we can apply the linear-pencil theory on $M_q$ over the random-matrix $X_0$ while considering the other random matrices as fixed. To this end, we note the constant part $C_q = \mean_{X_0}[M_q]$, and consider matrix of sub-traces $g \in \mathbb R^{7 \times 7}$ such that for squared-blocks $ij$, $g_{ij} = \traceLim[N_i]{(M_q^{-1})^{(i,j)}}$ where $N_i$ is the size of the block $ij$ in $M_q^{-1}$.
Then we apply the fixed-point equation described in Appendix D of \cite{bodin.2212.06757} with $g_{ ij } = \frac{1}{N_i} \trace{ ((C_q - \eta(g) \otimes I)^{-1})^{(ij)} }$ where $\eta(g)$ is the matrix defined by:
\begin{align}
    \eta(g) = \left(
        \begin{array}{ccccccc}
            0 & 0 & 0 & 0 & 0 & 0 & 0 \\
            0 & 0 & 0 & 0 & 0 & 0 & 0 \\
            0 & 0 & \psi g_{44} & 0 & 0 & \psi g_{44} & 0 \\
            0 & 0 & 0 & g_{33} + g_{36} +  & 0 & 0 & 0 \\
             &  &  & g_{63} + g_{66} &  &  &  \\
            0 & 0 & 0 & 0 & 0 & 0 & 0 \\
            0 & 0 & \psi g_{44} & 0 & 0 & \psi g_{44} & 0 \\
            0 & 0 & 0 & 0 & 0 & 0 & 0 
        \end{array}
    \right)
\end{align}
Further inversion of $C_q - \eta(g) \otimes I$ leads to:
\begin{align}
    g_{11} &= \traceLim[d]{ g_{44} \psi W_t (g_{44} \psi L_t + I_n)^{-1} W_t^T } \\
    g_{44} & = \frac{1}{1-g_{66}} \\
    g_{66} &= -\traceLim[n]{ L_t (g_{44} \psi L_t + I_n)^{-1}  }
\end{align}
Let $\Gamma \subset \mathbb C$ be a contour enclosing the eigenvalues of $H$, we use the fact that for any functional $f$ which applies on the eigenvalues of a matrix we have $f(H) = \frac{-1}{2\pi i} \oint_{\Gamma} f(z) (H-zI_n)^{-1} \dd z$ to obtain:
\begin{equation*}
    g_{11}  = \frac{-1}{2\pi i} \oint_{\Gamma} 
    \frac{g_{44} \psi e^{2zt}}{
        1 + g_{44} \psi \int_0^t 2 e^{2sz} \dd s
    } 
    \traceLim[d]{ (H - z I_n)^{-1} Z^* } \dd z 
\end{equation*}
which leads with $Q(z) = \traceLim[d]{ X^{*T} (H - z I_n)^{-1} X^* } $ to:
\begin{align}
    g_{11}  = \frac{-1}{2\pi i} \oint_{\Gamma} 
    \frac{g_{44} \psi z}{
        \psi g_{44} (1 -  e^{-2tz}) + z e^{-2tz}
    }  Q(z) \dd z 
\end{align}
Similarly with $P(z) = \traceLim[n]{ (H - z I_n)^{-1} } $
\begin{equation}
    g_{66} \psi = \frac{-1}{2\pi i} \oint_{\Gamma}
    \frac{
        1 - e^{-2tz}
    }{
        \psi g_{44} (1 -  e^{-2tz}) + z e^{-2tz}
    } P(z) \dd z
\end{equation}
We find the equations from the main results with $\tilde q_t = \frac{1}{\psi g_{44}}$.

\subsection{Tracking the norm $p_t$}
The term $p_t = \traceLim[d]{Z_t^2}$ can also be recovered from a similar calculation but would lead to design a much larger linear-pencil. Another method is to track directly the eigenvalues of $Z_t$ with the trace of the resolvent: $h_{11} = \traceLim[n]{ (Z_t-zI_n)^{-1} }$ with $h$ the solution of the fixed point equation (Appendix D in \cite{bodin.2212.06757}) stemming from the following linear-pencil:
\begin{equation}
    M_p = \left(
        \begin{array}{c||c|ccc|c}
            -zI_n & 0 & 0 & 0 & 0 & e^{tH}\\ \hline \hline
            0 & 0 & X_0 & 0 & 0 & I_n\\ \hline
            0 & X_0^T & I_m & 0 & X_0^T & 0\\
            0 & 0 & 0 & L_t & I_n & 0\\
            0 & 0 & X_0 & I_n & 0  & 0\\ \hline
            e^{tH} & I_n& 0 & 0 & 0 & 0
        \end{array}
    \right)
\end{equation}
Which yields the set of equations:
\begin{align*}
    h_{11} & = -\traceLim[n]{
        \left(L_t + \frac{1}{h_{33}} I_n\right)
        \left(e^{2tH} + z L_t + \frac{z}{h_{33}} I_n\right)^{-1}
    } \\
    h_{33} & =
    1 - 
    \frac{1}{\psi} \traceLim[n]{
        \left(z L_t + e^{2tH}\right)
        \left(e^{2tH} + z L_t + \frac{z}{h_{33}} I_n\right)^{-1}
    }
\end{align*}
Using the contour integration technique, we obtain:
\begin{equation}
    h_{11} = \frac{-1}{2\pi i} \oint_{\Gamma}
    -\frac{
        \frac{1}{h_{33}} + \int_0^t 2e^{2sx} \dd s
    }{
        \frac{z}{h_{33}} + e^{2tx} + z \int_0^t 2e^{2sx} \dd s
    } P(x) \dd x
\end{equation}
which is reduced to:
\begin{equation}
    h_{11}(z) = \frac{-1}{2\pi i} \oint_{\Gamma}
    -\frac{
        1 + e^{-2tx} (\frac{x}{h_{33}} - 1)
    }{
        x + z + z e^{-2tx} (\frac{x}{h_{33}} -1)
    } P(x) \dd x
\end{equation}
Similarly for $h_{33}$:
\begin{equation}
    h_{33}(z) = 1 + \frac{1}{\psi} \frac{-1}{2\pi i} \oint_{\Gamma}
    -\frac{
        (x + z - z e^{2tx}) P(x) \dd x
    }{
        x + z + z e^{-2tx} (\frac{x}{h_{33}} -1)
    } 
\end{equation}
Two possible ways to retrieve $p_t$ from $h_{11}$ and $h_{33}$: either with $\phi p_t = \frac{-1}{2\pi i} \oint_\Gamma z^2 h_{11}(z) \dd z$, or $\phi p_t =  - \frac12 \frac{\partial^{(2)}}{\partial z^2} \left( \frac{1}{z} h_{11}(\frac{1}{z}) \right) \rvert_{z=0}$.  In both cases, there is an additional level of complexity in terms of calculation as it either requires a double-contour integration, or computing derivative and second derivative of the given functions yielding further new equations.

\subsection{Quantities $Q(z),P(z)$}
There remains to calculate the terms $Q(z),P(z)$ which depends only on the random matrices $X^*, \xi$ and can be done altogether with the linear-pencil:
\begin{equation}
    M_{z} = \left(
        \begin{matrix}
            I_n & X^* & 0 & 0\\
            0 & I_d & X^{*T} & 0\\
            0 & 0 & (z + \mu) I_n- \frac{1}{\sqrt \lambda} \xi & X^*\\
            0 & 0 & X^{*T} & I_d
        \end{matrix}
    \right)
\end{equation}
Using the kernel $K = (H-zI_n)^{-1}$, we can calculate the inverse:
\begin{equation}
    M_z^{-1} = \left(
        \begin{matrix}
            I_n & -X^* & -Z^*K & Z^* K X^*\\
            0 & I_d & X^{*T} K & -X^{*T} K X^*\\
            0 & 0 & -K & K X^*\\
            0 & 0 & -X^{*T} K & I_d - X^{*T} K X^*
        \end{matrix}
    \right)    
\end{equation}
So that $Q(z) = -f_{13}$ and $P(z) = f_{33}$ where we $f$ is the analog of $g$ and $h$ with the former linear-pencils. In particular we expect the following structure:
\begin{equation}
    f = \left(
        \begin{matrix}
            1 & 0 & -\phi Q(z) & 0 \\
            0 & 1 & 0 & - Q(z) \\
            0 & 0 & -P(z) & 0 \\
            0 & 0 & 0 & 1-Q(z)
        \end{matrix}
    \right)
\end{equation}
We can further compute the fixed point equation with:
\begin{equation}
    \eta(f) = \left(
        \begin{matrix}
            0 & 0 & f_{22} \phi + f_{24} \phi & 0\\
            0 & f_{31} & 0 & f_{33}\\
            0 & 0 & \frac{f_{33}}{\lambda} + f_{42} \phi + f_{44} \phi & 0\\
            0 & f_{31} & 0 & f_{33}
        \end{matrix}
    \right)
\end{equation}
After some algebraic reductions, we obtain the degree 3 polynomials given in equation \eqref{eq:Q_and_P}. In general, these equations have multiple solutions but only one corresponds to the analytic solution associated to the appropriate trace of resolvent.

\section{Conclusion}
Our work primarily shows how we can take advantage of random matrix techniques to derive fixed-point equations solving the time evolution of the matrix-mean-square-error in the high-dimensional limit. Although we choose a specific data model, as future considerations, the matrix $H$ can be generalized to other structures for which the same methods would apply. In particular, if only the noise structure changes, then only $\rho_Q$ and $\rho_P$ are changed. We will come back to these issues in a more extensive and detailed contribution.

\vskip 0.25cm
{\bf Acknowledgments} The work of A. B is supported by Swiss National Fondation Grant no 200020 182517. We also acknowledge instructive discussions with Farzad Pourkamali and Jean Barbier.

\bibliographystyle{IEEEtran}
\bibliography{IEEEabrv,bibliography}

\begin{thebibliography}{10}
\providecommand{\url}[1]{#1}
\csname url@samestyle\endcsname
\providecommand{\newblock}{\relax}
\providecommand{\bibinfo}[2]{#2}
\providecommand{\BIBentrySTDinterwordspacing}{\spaceskip=0pt\relax}
\providecommand{\BIBentryALTinterwordstretchfactor}{4}
\providecommand{\BIBentryALTinterwordspacing}{\spaceskip=\fontdimen2\font plus
\BIBentryALTinterwordstretchfactor\fontdimen3\font minus
  \fontdimen4\font\relax}
\providecommand{\BIBforeignlanguage}[2]{{%
\expandafter\ifx\csname l@#1\endcsname\relax
\typeout{** WARNING: IEEEtran.bst: No hyphenation pattern has been}%
\typeout{** loaded for the language `#1'. Using the pattern for}%
\typeout{** the default language instead.}%
\else
\language=\csname l@#1\endcsname
\fi
#2}}
\providecommand{\BIBdecl}{\relax}
\BIBdecl

\bibitem{ChenChi}
Y.~Chen and Y.~Chi, ``Harnessing structures in big data via guaranteed low-rank
  matrix estimation: Recent theory and fast algorithms via convex and nonconvex
  optimization,'' \emph{IEEE Signal Processing Magazine}, vol.~35, no.~4, pp.
  14--31, 2018.

\bibitem{chi2019nonconvex}
Y.~Chi, Y.~M. Lu, and Y.~Chen, ``Nonconvex optimization meets low-rank matrix
  factorization: An overview,'' \emph{IEEE Transactions on Signal Processing},
  vol.~67, no.~20, pp. 5239--5269, 2019.

\bibitem{spectra}
R.~{Rashidi Far}, T.~{Oraby}, W.~{Bryc}, and R.~{Speicher}, ``{Spectra of large
  block matrices},'' \emph{arXiv e-prints}, p. cs/0610045, Oct. 2006.

\bibitem{mingo2017free}
J.~A. Mingo and R.~Speicher, \emph{Free probability and random matrices}.\hskip
  1em plus 0.5em minus 0.4em\relax Springer, 2017, vol.~35.

\bibitem{adlam2020neural}
\BIBentryALTinterwordspacing
B.~Adlam and J.~Pennington, ``The neural tangent kernel in high dimensions:
  Triple descent and a multi-scale theory of generalization,'' in
  \emph{Proceedings of the 37th International Conference on Machine Learning},
  ser. Proceedings of Machine Learning Research, H.~D. III and A.~Singh, Eds.,
  vol. 119.\hskip 1em plus 0.5em minus 0.4em\relax PMLR, 13--18 Jul 2020, pp.
  74--84. [Online]. Available:
  \url{http://proceedings.mlr.press/v119/adlam20a.html}
\BIBentrySTDinterwordspacing

\bibitem{bodin2021model}
A.~Bodin and N.~Macris, ``Model, sample, and epoch-wise descents: exact
  solution of gradient flow in the random feature model,'' \emph{Advances in
  Neural Information Processing Systems}, vol.~34, 2021.

\bibitem{bodin.2212.06757}
\BIBentryALTinterwordspacing
------, ``Gradient flow in the gaussian covariate model: exact solution of
  learning curves and multiple descent structures,'' 2022. [Online]. Available:
  \url{https://arxiv.org/abs/2212.06757}
\BIBentrySTDinterwordspacing

\bibitem{schwartz1958linear}
N.~Dunford and J.~T. Schwartz, \emph{Linear Operators}.\hskip 1em plus 0.5em
  minus 0.4em\relax Wiley Classics Library, 1988.

\bibitem{pmlr-v134-bodin21a}
\BIBentryALTinterwordspacing
A.~Bodin and N.~Macris, ``Rank-one matrix estimation: analytic time evolution
  of gradient descent dynamics,'' in \emph{Proceedings of Thirty Fourth
  Conference on Learning Theory}, ser. Proceedings of Machine Learning
  Research, M.~Belkin and S.~Kpotufe, Eds., vol. 134.\hskip 1em plus 0.5em
  minus 0.4em\relax PMLR, 15--19 Aug 2021, pp. 635--678. [Online]. Available:
  \url{https://proceedings.mlr.press/v134/bodin21a.html}
\BIBentrySTDinterwordspacing

\bibitem{tarmoun2021understanding}
S.~Tarmoun, G.~Franca, B.~D. Haeffele, and R.~Vidal, ``Understanding the
  dynamics of gradient flow in overparameterized linear models,'' in
  \emph{International Conference on Machine Learning}.\hskip 1em plus 0.5em
  minus 0.4em\relax PMLR, 2021, pp. 10\,153--10\,161.

\bibitem{gunasekar2017implicit}
S.~Gunasekar, B.~E. Woodworth, S.~Bhojanapalli, B.~Neyshabur, and N.~Srebro,
  ``Implicit regularization in matrix factorization,'' \emph{Advances in Neural
  Information Processing Systems}, vol.~30, 2017.

\bibitem{chou2020gradient}
H.-H. Chou, C.~Gieshoff, J.~Maly, and H.~Rauhut, ``Gradient descent for deep
  matrix factorization: Dynamics and implicit bias towards low rank,''
  \emph{arXiv preprint arXiv:2011.13772}, 2020.

\bibitem{saxe2013exact}
A.~M. Saxe, J.~L. McClelland, and S.~Ganguli, ``Exact solutions to the
  nonlinear dynamics of learning in deep linear neural networks,'' \emph{arXiv
  preprint arXiv:1312.6120}, 2013.

\bibitem{mannelli2019passed}
\BIBentryALTinterwordspacing
S.~S. Mannelli, F.~Krzakala, P.~Urbani, and L.~Zdeborova, ``Passed and
  spurious: Descent algorithms and local minima in spiked matrix-tensor
  models,'' ser. Proceedings of Machine Learning Research, K.~Chaudhuri and
  R.~Salakhutdinov, Eds., vol.~97.\hskip 1em plus 0.5em minus 0.4em\relax Long
  Beach, California, USA: PMLR, 09--15 Jun 2019, pp. 4333--4342. [Online].
  Available: \url{http://proceedings.mlr.press/v97/mannelli19a.html}
\BIBentrySTDinterwordspacing

\bibitem{BenArous-et-al-2022}
\BIBentryALTinterwordspacing
G.~B. Arous, R.~Gheissari, and A.~Jagannath, ``High-dimensional limit theorems
  for sgd: Effective dynamics and critical scaling,'' 2022. [Online].
  Available: \url{https://arxiv.org/abs/2206.04030}
\BIBentrySTDinterwordspacing

\bibitem{Liang-Sen-Sur-2022}
T.~Liang, S.~Sen, and P.~Sur, ``High-dimensional asymptotics of langevin
  dynamics in spiked matrix models,'' 2022.

\bibitem{Montanari-Richard-2014}
A.~Montanari and E.~Richard, ``A statistical model for tensor pca,'' in
  \emph{Proceedings of the 27th International Conference on Neural Information
  Processing Systems - Volume 2}, ser. NIPS 2014.\hskip 1em plus 0.5em minus
  0.4em\relax Cambridge, MA, USA: MIT Press, 2014, pp. 2897--2905.

\bibitem{lelarge2019fundamental}
M.~Lelarge and L.~Miolane, ``Fundamental limits of symmetric low-rank matrix
  estimation,'' \emph{Probability Theory and Related Fields}, vol. 173, no.~3,
  pp. 859--929, 2019.

\bibitem{luneau2020high}
C.~Luneau, N.~Macris, and J.~Barbier, ``High-dimensional rank-one nonsymmetric
  matrix decomposition: the spherical case,'' in \emph{2020 IEEE International
  Symposium on Information Theory (ISIT)}.\hskip 1em plus 0.5em minus
  0.4em\relax IEEE, 2020, pp. 2646--2651.

\bibitem{barbier2019adaptive}
J.~Barbier and N.~Macris, ``The adaptive interpolation method: a simple scheme
  to prove replica formulas in {B}ayesian inference,'' \emph{Probability theory
  and related fields}, vol. 174, no.~3, pp. 1133--1185, 2019.

\bibitem{miolane2017fundamental}
L.~Miolane, ``Fundamental limits of low-rank matrix estimation: the
  non-symmetric case,'' \emph{arXiv preprint arXiv:1702.00473}, 2017.

\bibitem{pourkamali2021mismatched}
F.~Pourkamali and N.~Macris, ``Mismatched estimation of symmetric rank-one
  matrices under gaussian noise,'' in \emph{International Zurich Seminar on
  Information and Communication (IZS 2022). Proceedings}.\hskip 1em plus 0.5em
  minus 0.4em\relax ETH Zurich, 2022, pp. 84--88.

\bibitem{pourkamali2022mismatched}
------, ``Mismatched estimation of non-symmetric rank-one matrices under
  {G}aussian noise,'' in \emph{2022 IEEE International Symposium on Information
  Theory (ISIT)}.\hskip 1em plus 0.5em minus 0.4em\relax IEEE, 2022, pp.
  1288--1293.

\bibitem{CamilliContucciMingione}
\BIBentryALTinterwordspacing
F.~Camilli, P.~Contucci, and E.~Mingione, ``{An inference problem in a
  mismatched setting: a spin-glass model with Mattis interaction},''
  \emph{SciPost Phys.}, vol.~12, p. 125, 2022. [Online]. Available:
  \url{https://scipost.org/10.21468/SciPostPhys.12.4.125}
\BIBentrySTDinterwordspacing

\bibitem{barbier2022price}
J.~Barbier, T.~Hou, M.~Mondelli, and M.~S{\'a}enz, ``The price of ignorance:
  how much does it cost to forget noise structure in low-rank matrix
  estimation?'' \emph{arXiv preprint arXiv:2205.10009}, 2022.

\bibitem{DBLP:conf/isit/LesieurMLKZ17}
\BIBentryALTinterwordspacing
T.~Lesieur, L.~Miolane, M.~Lelarge, F.~Krzakala, and L.~Zdeborov{\'{a}},
  ``Statistical and computational phase transitions in spiked tensor
  estimation,'' in \emph{2017 {IEEE} International Symposium on Information
  Theory, {ISIT} 2017, Aachen, Germany, June 25-30, 2017}.\hskip 1em plus 0.5em
  minus 0.4em\relax {IEEE}, 2017, pp. 511--515. [Online]. Available:
  \url{https://doi.org/10.1109/ISIT.2017.8006580}
\BIBentrySTDinterwordspacing

\bibitem{Lesieur_2017}
\BIBentryALTinterwordspacing
T.~Lesieur, F.~Krzakala, and L.~Zdeborová, ``Constrained low-rank matrix
  estimation: phase transitions, approximate message passing and
  applications,'' \emph{Journal of Statistical Mechanics: Theory and
  Experiment}, vol. 2017, no.~7, p. 073403, jul 2017. [Online]. Available:
  \url{https://dx.doi.org/10.1088/1742-5468/aa7284}
\BIBentrySTDinterwordspacing

\bibitem{Montanari2017EstimationOL}
A.~Montanari and R.~Venkataramanan, ``Estimation of low-rank matrices via
  approximate message passing,'' \emph{The Annals of Statistics}, 2017.

\bibitem{Kabashima_2016}
\BIBentryALTinterwordspacing
Y.~Kabashima, F.~Krzakala, M.~Mezard, A.~Sakata, and L.~Zdeborova, ``Phase
  transitions and sample complexity in bayes-optimal matrix factorization,''
  \emph{{IEEE} Transactions on Information Theory}, vol.~62, no.~7, pp.
  4228--4265, jul 2016. [Online]. Available:
  \url{https://doi.org/10.1109%2Ftit.2016.2556702}
\BIBentrySTDinterwordspacing

\bibitem{barbier2021statistical}
J.~Barbier and N.~Macris, ``Statistical limits of dictionary learning: random
  matrix theory and the spectral replica method,'' \emph{Physical Review E},
  vol. 106, no.~2, p. 024136, 2022.

\bibitem{maillard2021perturbative}
A.~Maillard, F.~Krzakala, M.~M{\'e}zard, and L.~Zdeborov{\'a}, ``Perturbative
  construction of mean-field equations in extensive-rank matrix factorization
  and denoising,'' \emph{Journal of Statistical Mechanics: Theory and
  Experiment}, vol. 2022, no.~8, p. 083301, 2022.

\bibitem{troiani2022optimal}
E.~Troiani, V.~Erba, F.~Krzakala, A.~Maillard, and L.~Zdeborov{\'a}, ``Optimal
  denoising of rotationally invariant rectangular matrices,'' \emph{arXiv
  preprint arXiv:2203.07752}, 2022.

\bibitem{camilli2022matrix}
F.~Camilli and M.~M{\'e}zard, ``Matrix factorization with neural networks,''
  \emph{arXiv preprint arXiv:2212.02105}, 2022.

\bibitem{baik2005phase}
J.~Baik, G.~B. Arous, and S.~P{\'e}ch{\'e}, ``Phase transition of the largest
  eigenvalue for nonnull complex sample covariance matrices,'' \emph{Annals of
  Probability}, p. 1643, 2005.

\bibitem{Pch2004TheLE}
S.~P{\'e}ch{\'e}, ``The largest eigenvalue of small rank perturbations of
  hermitian random matrices,'' \emph{Probability Theory and Related Fields},
  vol. 134, pp. 127--173, 2004.

\bibitem{benaych2011eigenvalues}
F.~Benaych-Georges and R.~R. Nadakuditi, ``The eigenvalues and eigenvectors of
  finite, low rank perturbations of large random matrices,'' \emph{Advances in
  Mathematics}, vol. 227, no.~1, pp. 494--521, 2011.

\bibitem{bun2017cleaning}
J.~Bun, J.-P. Bouchaud, and M.~Potters, ``Cleaning large correlation matrices:
  tools from random matrix theory,'' \emph{Physics Reports}, vol. 666, pp.
  1--109, 2017.

\bibitem{marchenko1967distribution}
V.~A. Marchenko and L.~A. Pastur, ``Distribution of eigenvalues for some sets
  of random matrices,'' \emph{Matematicheskii Sbornik}, vol. 114, no.~4, pp.
  507--536, 1967.

\end{thebibliography}


\end{document}